\documentclass[11pt]{article}

\usepackage[final]{acl}

\usepackage{times}
\usepackage{latexsym}

\usepackage[T1]{fontenc}

\usepackage[utf8]{inputenc}

\usepackage{microtype}

\usepackage{inconsolata}

\usepackage{graphicx}
\usepackage{amsmath}
\usepackage{booktabs}
\usepackage{multirow}
\usepackage[most]{tcolorbox}

\title{CueMem: Cue-Guided Context Reconstruction for Long-Term Conversational Memory}

\author{
 \textbf{Changjian Wang\textsuperscript{1,2}},
 \textbf{Rongzhen Li\textsuperscript{1}},
 \textbf{Weili Guan\textsuperscript{2}},
 \textbf{Shuming Shi\textsuperscript{1}},
 \textbf{Quan Lu\textsuperscript{1}},
 \textbf{Ning Jiang\textsuperscript{1}}
\\
\textsuperscript{1}Mashang Consumer Finance Co., Ltd. \\
\textsuperscript{2}Harbin Institute of Technology, Shenzhen
\\
 \small{
   \texttt{changjian.wang@msxf.com}
 }
}

\begin{document}
\maketitle
\begin{abstract}
Long-term conversational agents must answer user queries by recalling information from extended dialogue histories, yet directly using the full history is costly and often unreliable, while compressed memory units may lose fine-grained evidence needed for question answering. Motivated by the reconstructive view of autobiographical memory, we propose CueMem, a cue-guided framework that treats extracted memory records as retrieval cues rather than self-contained evidence and reconstructs query-relevant dialogue context from their source turns. During memory construction, CueMem extracts fine-grained memory cues from dialogue turns and links each cue to its source turn. At query time, it retrieves query-relevant cues, maps them to source-turn anchors, and expands from these anchors over a turn graph that captures temporal proximity and semantic relatedness, reconstructing a compact evidence context from the original dialogue for LLM answer generation. Experiments on LoCoMo and LongMemEval show that CueMem consistently outperforms representative long-term memory baselines. Further analyses show that graph-based context reconstruction helps recover supporting dialogue evidence while reducing query-time input tokens and latency compared with the full-history LLM setting. These results highlight retrieval cues as an effective alternative to self-contained memory evidence for long-term conversational question answering.
\end{abstract}

\section{Introduction}

\begin{figure}[t]
  \centering
  \includegraphics[width=\linewidth]{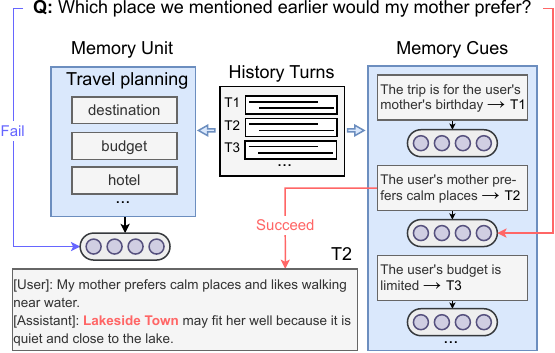}
  \caption{An example comparison between memory units and memory cues. A coarse memory unit may lose fine-grained semantic information due to compression or mixed semantic representations, e.g., a ``travel planning'' memory unit may merge destination, budget, and hotel preferences, making it difficult to obtain the evidence for the user's query. In contrast, fine-grained memory cues linked to source turns provide focused retrieval targets, enabling the system to recover the relevant source turn T2 and reconstruct the dialogue context needed to answer \textit{Lakeside Town}.}
  \label{fig:mtv}
\end{figure}

Large Language Models (LLMs) have developed rapidly in recent years~\citep{achiam2023gpt,yang2025qwen3}, leading to substantial improvements in the capabilities of LLM-powered conversational agents. In real-world scenarios, conversations are often long-running and multi-turn, requiring agents to answer user queries by effectively leveraging long-term dialogue history. Naively incorporating the full interaction history is often ineffective due to context length limitations, noisy information, the ``lost-in-the-middle'' phenomenon~\citep{liu2024lost}, and increased inference latency. Therefore, how to model and utilize long-term conversational memory has become a major challenge for conversational agents.

Existing methods typically construct retrievable memory units from dialogue history through segmentation, compression, summarization, or salient information extraction, and retrieve the most relevant units as context for generation~\citep{zhong2024memorybank,pan2025secom,fang2026lightmem}. However, such memory units often remain compressed or decontextualized fragments of the original dialogue, and vector-based indexing may entangle multiple semantic signals within a single dense representation. More importantly, many existing systems treat retrieved memory units as self-contained evidence for generation. This design contrasts with the reconstructive view of autobiographical memory, which suggests that remembering is guided by cues and reconstructed from broader autobiographical knowledge stores~\citep{conway1993structure,conway2000construction,holland2010emotion}.

For example, as shown in Figure~\ref{fig:mtv}, an existing memory system may group several turns about destination, budget, and hotel preference into a single memory unit about ``travel planning''. However, a later query such as ``Which place we mentioned earlier would my mother prefer?'' targets only a fine-grained aspect of this topic, making retrieval challenging because the relevant signal is mixed with multiple other semantics within the same retrievable unit. In contrast, fine-grained cues such as ``the user's mother prefers calm places'' provide more focused retrieval targets and can guide the system back to the original dialogue turns for context reconstruction. Inspired by this cognitive perspective, we argue that long-term conversational memory should follow a similar retrieval-and-reconstruction process: first identifying fine-grained cues that match the current query, and then recovering the original dialogue context around these cues before generating the answer.

In this paper, we propose CueMem, a simple and effective long-term memory system for agent conversations, inspired by the reconstructive view of autobiographical memory. CueMem first constructs fine-grained memory cues from dialogue turns, with each cue serving as a semantically focused retrieval target linked to its source turn. Given a user query, CueMem retrieves the most relevant cues and uses their source turns as anchors for context reconstruction. To recover supporting evidence, CueMem builds a turn graph that connects dialogue turns based on temporal proximity and semantic similarity, and expands from the anchored turns over this graph to collect query-relevant dialogue context. Finally, the reconstructed context is provided to the LLM as memory evidence for answer generation. We evaluate CueMem on two long-term conversational question answering benchmarks, LoCoMo~\citep{maharana2024evaluating} and LongMemEval~\citep{wu2025longmemeval}, and show that it consistently outperforms baseline memory models. These results demonstrate the effectiveness of retrieving fine-grained memory cues and reconstructing dialogue context from anchored turns.

Our contributions are summarized as follows:
\begin{itemize}
  \item We introduce a cue-centered view of long-term conversational memory, in which fine-grained memory cues guide the reconstruction of relevant dialogue context rather than serving as self-contained evidence for generation.

  \item We propose CueMem, a simple yet effective framework that constructs fine-grained memory cues from dialogue turns, retrieves query-relevant cues, and recovers supporting dialogue context by expanding from their source-turn anchors on a turn-level graph.

  \item Experiments on two long-term conversational question answering benchmarks show that CueMem consistently outperforms strong long-term memory baselines, demonstrating the effectiveness of fine-grained cue retrieval and anchor-based context reconstruction.
\end{itemize}

\section{Related Work}
Retrieval-Augmented Generation (RAG) augments language models by retrieving relevant external context before generation~\citep{lewis2020retrieval,gao2023retrieval}. In long-context and conversational settings, RAG is commonly used to retrieve dialogue turns, chunks, summaries, or memory entries, reducing the need to process the full history and alleviating long-context issues such as the ``lost-in-the-middle'' problem~\citep{liu2024lost}. Recent work also explores more structured retrieval mechanisms, such as graph-based RAG~\citep{edge2024local,guo2024lightrag,gutierrez2024hipporag,gutierrez2025rag}, which organizes documents or entities into graphs to support multi-hop or global information access. However, these RAG systems mainly target static external knowledge bases, and their fixed chunking and retrieval can suffer from granularity mismatch in dynamic long-term conversations.

Long-term conversational memory aims to help LLM agents retain and use information from extended interactions. Early agent memory systems store observations or dialogue histories as memory streams and retrieve relevant records for future reasoning or response generation~\citep{park2023generative,zhong2024memorybank,packer2023memgpt}. More recent methods further organize memories through hierarchical modules, structured notes, graph links, or consolidation mechanisms to support scalable storage, updating, and retrieval~\citep{kang2025memory,chhikara2025mem0,xu2026mem,fang2026lightmem}. However, these methods often rely on compressed summaries or memory notes as the direct evidence for generation, which may lose fine-grained details and local dialogue context. Moreover, merging multiple facts into one memory unit can entangle different semantic signals and reduce retrieval precision.

Cognitive studies of autobiographical memory provide a useful perspective for addressing this issue. Autobiographical memory is commonly defined as ``memory for the events of one's life''~\citep{conway1993structure}. Although agent memory is not human memory, long-term conversational memory plays a functionally analogous role: it records events from an agent's interaction history and supports later recall in response to user queries. Cognitive studies suggest that autobiographical memories are not stored as perfect records, but are reconstructed from broader autobiographical knowledge with the help of cues~\citep{holland2010emotion,conway2000construction}. This perspective motivates CueMem, which treats stored memories as fine-grained retrieval cues for reconstructing evidence rather than as self-contained evidence.

\section{Problem Formulation}

We focus on long-term conversational question answering, where an agent answers a user query given a long dialogue history. Formally, let $\mathcal{D} = \{u_1, u_2, \ldots, u_T\}$ denote a dialogue history consisting of $T$ turns. The dialogue history may span multiple sessions, where each session is a temporally contiguous block of turns, but we represent the entire history as a chronological sequence of turns. Each turn $u_i$ consists of a user utterance and the corresponding agent response, and may include associated metadata such as timestamp. Given a user query $q$, the goal is to generate an answer $a$ that is both relevant to $q$ and faithful to the information contained in $\mathcal{D}$. Since $\mathcal{D}$ can be long, noisy, and only partially relevant to the current query, the system should identify a compact evidence context rather than rely on the entire dialogue history as input. We formulate the task as selecting a query-relevant evidence context $\mathrm{E}_q \subseteq \mathcal{D}$ and generating the answer conditioned on both the query and this context:
\[
  a = \mathrm{LLM}(\mathrm{Prompt}(q, \mathrm{E}_q)),
\]
where $\mathrm{E}_q$ contains the historical information necessary for answering $q$, and $\mathrm{Prompt}(\cdot)$ denotes the answer-generation prompt that formats the query and evidence context for the LLM.

\begin{figure*}[t]
  \centering
  \includegraphics[width=\textwidth]{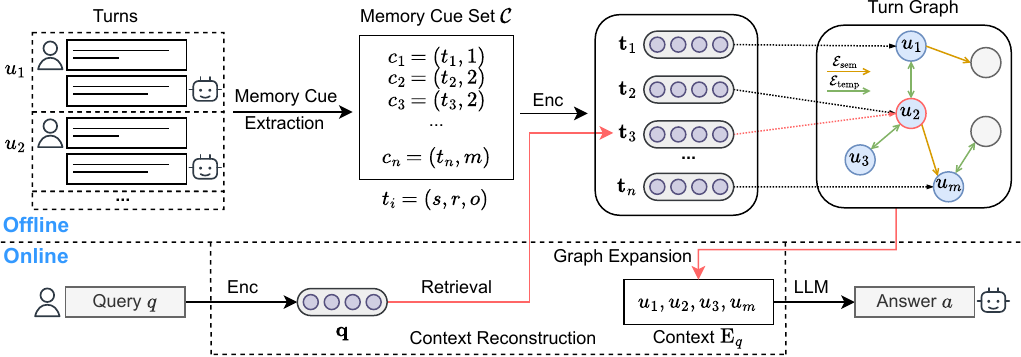}
  \caption{Overview of CueMem. In the offline stage, CueMem extracts fine-grained memory cues from dialogue turns, links each cue to its source turn, encodes the cues into dense vectors for retrieval, and builds a turn graph with temporal and semantic edges. In the online stage, CueMem encodes the user query into the same vector space, retrieves query-relevant cues, maps them to source-turn anchors, expands over the turn graph to reconstruct supporting dialogue context, and feeds the reconstructed context to the LLM for answer generation.}
  \label{fig:arc}
\end{figure*}

\section{Methodology}

CueMem follows a cue-to-anchor-to-context memory reconstruction pipeline, as illustrated in Figure~\ref{fig:arc}. Given a long dialogue history, CueMem first extracts fine-grained memory cues from individual dialogue turns and links each cue to its source turn. It then builds a turn-level graph over the original dialogue history, where temporal edges preserve local conversational continuity and semantic edges connect turns based on cue-level semantic similarity. At query time, CueMem retrieves query-relevant cues, maps them to source-turn anchors, and expands from these anchors on the turn graph to reconstruct the evidence context. The reconstructed context is then provided to the LLM for answer generation.

\subsection{Memory Cue Extraction}
The first step of CueMem is to convert the dialogue history into a set of fine-grained memory cues. Unlike segment- or summary-level memories that compress multiple semantic signals into a single unit, each cue is designed to capture a relatively atomic piece of information from the dialogue. Memory cues can take various forms, such as natural language statements or structured relational records. For simplicity and interpretability, we instantiate each cue as a relational triple in this work. Formally, given the dialogue history $\mathcal{D}$, we extract a cue set $\mathcal{C} = \{c_1, c_2, \ldots, c_N\}$. Each cue is represented as $c_i = (t_i, \tau_i)$, where $t_i$ denotes a triple $(\textit{subject}, \textit{relation}, \textit{object})$ extracted from the dialogue and $\tau_i \in \{1, 2, \ldots, T\}$ denotes a pointer to the source turn in $\mathcal{D}$. The triple provides a semantically focused representation for retrieval, while the pointer links the cue back to the original dialogue turn for later context reconstruction. In practice, we use an LLM to extract relational triples from each dialogue turn to construct memory cues. A turn may yield multiple cues if it contains several useful pieces of information, or no cue if it does not contain any. We further encode each triple $t_i$ with an embedding model to obtain its dense representation $\mathbf{t}_i = \mathrm{Enc}(t_i)$, which is used for subsequent retrieval and graph construction.

\subsection{Turn Graph Construction}
After extracting memory cues, CueMem builds a turn-level graph over the original dialogue history to support later context reconstruction. The graph preserves dialogue structure beyond isolated cue matches, allowing the system to recover temporally and semantically related turns once a source turn is selected as an anchor. Formally, we construct a graph $\mathcal{G} = (\mathcal{V}, \mathcal{E})$, where each node $v_i \in \mathcal{V}$ corresponds to a dialogue turn $u_i \in \mathcal{D}$. The edge set $\mathcal{E}$ consists of two types of edges: temporal edges $\mathcal{E}_{\mathrm{temp}}$ and semantic edges $\mathcal{E}_{\mathrm{sem}}$. Temporal edges capture chronological continuity between dialogue turns, while semantic edges capture semantic associations between turns. We describe the construction of these two edge types below.

\paragraph{Temporal edges.}
Temporal edges capture temporal proximity between dialogue turns. For each turn node $v_i$, we define its temporal neighbors as the turns within a fixed chronological window:
\[
  \mathcal{N}_{\mathrm{temp}}(v_i)
  =
  \{v_j \in \mathcal{V} \mid 0 < |i-j| \leq w\},
\]
where $w$ is the window size. For each $v_j \in \mathcal{N}_{\mathrm{temp}}(v_i)$, we add a directed temporal edge:
\[
  (v_i, v_j) \in \mathcal{E}_{\mathrm{temp}}.
\]
These edges help recover local context around an anchored turn, including turns that may be difficult to retrieve directly but are useful for the LLM to understand the conversational context and generate a faithful answer.

\paragraph{Semantic edges.}
Semantic edges are used to capture non-local relations between turns that are semantically related but may be far apart in the dialogue history. We construct these semantic edges by computing similarity at the cue level and then mapping similar cues back to their source turns. Compared with turn-level similarity, cue-level similarity relies on more atomic semantic representations, reducing the interference of mixed semantics and making the resulting associations more focused and precise. More specifically, for each cue $c_i = (t_i, \tau_i)$, we retrieve its top-$k$ nearest cues according to cosine similarity between cue embeddings:
\[
  \mathcal{N}_{\mathrm{sem}}(c_i)
  =
  \mathrm{TopK}_{c_j \in \mathcal{C} \setminus \{c_i\}}
  \cos(\mathbf{t}_i, \mathbf{t}_j).
\]
For each retrieved cue $c_j = (t_j, \tau_j) \in \mathcal{N}_{\mathrm{sem}}(c_i)$, we map both cues to their source-turn nodes and add a directed semantic edge from $v_{\tau_i}$ to $v_{\tau_j}$ in the turn graph:
\[
  (v_{\tau_i}, v_{\tau_j}) \in \mathcal{E}_{\mathrm{sem}}.
\]

\subsection{Context Reconstruction}
Given a user query, CueMem reconstructs the evidence context through a cue-to-anchor-to-context process. It first retrieves fine-grained memory cues that are semantically relevant to the query. The source turns linked to these retrieved cues are then treated as anchors in the turn graph. Starting from these anchors, CueMem expands along temporal and semantic edges to recover both local conversational context and non-local semantically related evidence. The selected turns are finally assembled as the evidence context for answer generation.

\paragraph{Cue retrieval.}
We first encode the user query $q$ with the same embedding model used for memory cues, obtaining $\mathbf{q} = \mathrm{Enc}(q)$. CueMem then retrieves the top-$m$ cues that are most similar to the query:
\[
  \mathcal{C}_q =
  \mathrm{TopK}_{c_i \in \mathcal{C}}
  \cos(\mathbf{q}, \mathbf{t}_i),
\]
where $m$ is the number of retrieved cues. The source-turn pointers of these cues define the anchor set:
\[
  \mathcal{A}_q = \{\tau_i \mid c_i \in \mathcal{C}_q\}.
\]

\paragraph{Graph expansion.}
Starting from the anchor set $\mathcal{A}_q$, CueMem expands over the turn graph $\mathcal{G}$ to collect supporting turns connected by temporal and semantic edges. We first collect the one-hop graph neighbors of the anchors:
\[
  \mathcal{N}_{\mathcal{G}}(\mathcal{A}_q)
  =
  \{j \mid \exists i \in \mathcal{A}_q,\ (v_i, v_j) \in \mathcal{E}_{\mathrm{temp}} \cup \mathcal{E}_{\mathrm{sem}}\}.
\]
The reconstructed evidence context is then formed by the anchor turns and their graph neighbors:
\[
  \mathrm{E}_q =
  \{u_j \mid j \in \mathcal{A}_q \cup \mathcal{N}_{\mathcal{G}}(\mathcal{A}_q)\}.
\]
This expansion collects turns within a limited graph neighborhood around the anchors, so that the evidence context includes both local dialogue context and semantically related non-local turns while avoiding excessive irrelevant history. The selected turns are finally ordered according to their original chronological positions before being passed to the LLM.

\subsection{Memory Management}
CueMem manages memory through a lightweight design based on cue records, source-turn pointers, and graph edges. This design separates offline construction from online reconstruction and supports simple memory operations.

\paragraph{Online and offline workflow.}
CueMem separates offline memory construction from online query-time reconstruction and answering. Memory cue extraction and turn graph construction are conducted offline, whereas cue retrieval, context reconstruction, and answer generation are performed online for each incoming query. This workflow avoids processing the full dialogue history during inference, while still enabling the system to recover rich dialogue evidence through cue-guided graph expansion. Moreover, since memory cues contain relatively atomic semantic information, they can be effectively represented by lightweight embedding models, allowing CueMem to balance retrieval efficiency and quality.

\paragraph{Memory maintenance.}
CueMem avoids complex memory maintenance operations by operating on simple cue records, source-turn pointers, and graph edges, rather than managing consolidated memory units. For \textsc{ADD}, CueMem extracts cues from newly observed dialogue turns and inserts the corresponding cue records and source-turn nodes in the background. Since cues are constructed at the turn level, new memories can be added in near real time, without waiting for a large amount of dialogue history to accumulate for segmentation or compression. For \textsc{DELETE}, CueMem can adopt standard forgetting policies, such as time-based decay or least-recently-used removal, to delete obsolete turn nodes and their associated cue records and graph edges. For \textsc{UPDATE}, we do not introduce a specialized updating mechanism, instead, updated information is added as new cue records linked to newer source turns. During generation, the model is instructed to prioritize more recent evidence. The effectiveness of this simple strategy is further examined in the ablation study.

\section{Experiments}

\subsection{Experimental Settings}
\paragraph{Datasets.}
We evaluate CueMem on two long-term conversational question answering benchmarks: LoCoMo~\citep{maharana2024evaluating} and LongMemEval~\citep{wu2025longmemeval}. For LoCoMo, we follow prior work~\citep{fang2026lightmem} and evaluate on 10 long conversations with 1,540 questions covering single-hop, multi-hop, temporal, and open-domain categories. For LongMemEval, we use the LongMemEval-S setting, which contains 500 evaluation questions spanning single-session user, assistant, and preference questions, as well as multi-session, temporal-reasoning, and knowledge-update questions. Detailed dataset statistics are provided in the Appendix.

\begin{table*}[t!]
\centering
\small
\setlength{\tabcolsep}{8pt}
\begin{tabular}{lccccc}
\toprule
\multirow{2}{*}{\textbf{Method}} & \multicolumn{4}{c}{\textbf{Category}} & \multirow{2}{*}{\textbf{Overall}} \\
\cmidrule(lr){2-5}
 & \textbf{Single-hop} & \textbf{Multi-hop} & \textbf{Temporal} & \textbf{Open-domain} &  \\
\midrule
Naive RAG & \underline{79.67} & 68.79 & 68.54 & 62.50 & 74.29 \\
MemoryOS~\citep{kang2025memory} & 75.51 & 73.76 & \underline{78.19} & \textbf{77.08} & \underline{75.84} \\
Mem0~\cite{chhikara2025mem0} & 46.61 & 43.97 & 38.32 & 50.00 & 44.61 \\
A-Mem~\citep{xu2026mem} & 77.76 & \underline{74.82} & 70.72 & 67.71 & 75.13 \\
LightMem~\citep{fang2026lightmem} & 76.58 & 72.70 & 66.67 & 53.13 & 72.34 \\
\midrule
\textbf{CueMem} & \textbf{84.07} & \textbf{75.18} & \textbf{82.24} & \underline{68.75} & \textbf{81.10} \\
\bottomrule
\end{tabular}
\caption{Accuracy results on the LoCoMo dataset.}
\label{tab:locomo-results}
\end{table*}

\begin{table*}[t!]
\centering
\small
\setlength{\tabcolsep}{3pt}
\begin{tabular}{lccccccc}
\toprule
\multirow{2}{*}{\textbf{Method}} & \multicolumn{6}{c}{\textbf{Category}} & \multirow{2}{*}{\textbf{Overall}} \\
\cmidrule(lr){2-7}
 & \textbf{Single-user} & \textbf{Single-assistant} & \textbf{Single-preference}  & \textbf{Multi-session} & \textbf{Temporal} & \textbf{Knowledge-update} &  \\
\midrule
Naive RAG  & 78.57 & \textbf{100.00} & 26.67 & 48.12 & 54.14 & 75.64 & 62.80 \\
MemoryOS & 88.57 & 62.50 & \underline{36.67} & \underline{60.90} & 57.89 & 74.36 & 64.80 \\
Mem0 & 74.29 & 12.50 & 20.00 & 45.86 & 49.62 & 71.79 & 49.60 \\
A-Mem & \textbf{94.29} & \underline{98.21} & 26.67 & 55.64 & 51.13 & 78.21 & 66.40 \\
LightMem & \underline{91.43} & 37.50 & 33.33 & \textbf{75.94} & \textbf{79.70} & \underline{84.62} & \underline{73.60} \\
\midrule
\textbf{CueMem} & \textbf{94.29} & 91.07 & \textbf{40.00} & \underline{60.90} & \underline{73.68} & \textbf{87.18} & \textbf{75.20} \\
\bottomrule
\end{tabular}
\caption{Accuracy results on LongMemEval dataset.}
\label{tab:longmemeval-results}
\end{table*}

\paragraph{Metrics.}
We report accuracy as the primary evaluation metric. Following prior work~\citep{wu2025longmemeval,fang2026lightmem}, we use an LLM-as-judge protocol to compare each generated answer with the reference answer given the corresponding question. The judge determines whether the prediction correctly answers the question, and accuracy is computed as the proportion of predictions judged correct.

\paragraph{Baselines.}
We compare CueMem with representative long-term memory systems for conversational agents:
\begin{itemize}
    \item \textbf{Naive RAG (turn-level)} retrieves dialogue turns directly from the full conversation history using semantic similarity and uses the retrieved turns as context for answer generation.
    \item \textbf{MemoryOS}~\citep{kang2025memory} organizes conversational memory with an OS-inspired hierarchical architecture, including short-term, mid-term, and long-term memory units with dynamic updating and retrieval.
    \item \textbf{Mem0}~\citep{chhikara2025mem0} builds scalable long-term memory for production-ready agents by extracting and maintaining memories from dialogue history through LLM-guided operations.
    \item \textbf{A-Mem}~\citep{xu2026mem} constructs structured memory notes and dynamically links related memories following the Zettelkasten principle, enabling agentic memory organization and evolution.
    \item \textbf{LightMem}~\citep{fang2026lightmem} filters, groups, and consolidates dialogue history through sensory, short-term, and long-term memory stages for lightweight memory-augmented generation.
\end{itemize}

\paragraph{Implementation details.}
For all methods, we use Llama-3.3-70B-Instruct as the same LLM backbone for memory extraction, answer generation, and LLM-as-judge evaluation. We adopt all-MiniLM-L6-v2 as a lightweight embedding model for encoding memory and user queries. Additional results with other LLM and embedding models, as well as more detailed experimental settings, are provided in the Appendix.

\begin{figure*}[t]
  \centering
  \includegraphics[width=\textwidth]{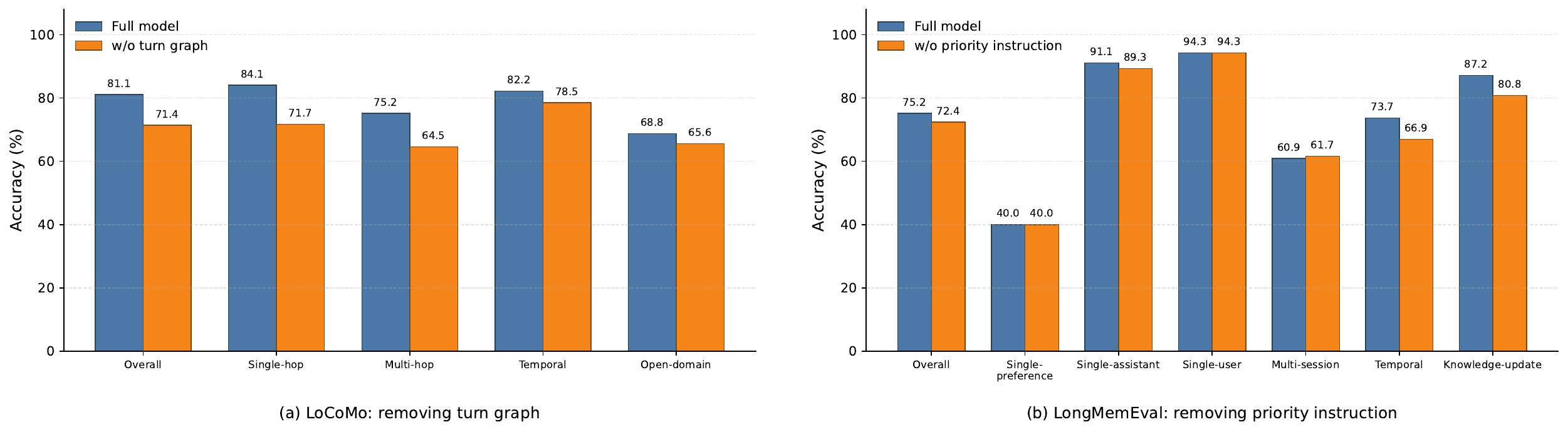}
  \caption{Ablation results on LoCoMo and LongMemEval.}
  \label{fig:ablation}
\end{figure*}

\subsection{Main Results}
We conduct experiments on LoCoMo and LongMemEval to evaluate the effectiveness of CueMem. Table~\ref{tab:locomo-results} shows the results on LoCoMo. CueMem achieves the best overall accuracy, outperforming all baseline methods, and surpassing the strongest baseline by a margin of 5.26\%. Analyzing performance across question types, CueMem shows clear advantages on single-hop and temporal questions, improving over the baseline by 4.40\% and 4.05\%, respectively. This suggests that fine-grained memory cues are effective for locating specific historical information, while graph-based expansion helps recover temporally related evidence around the retrieved anchors. CueMem also performs well on multi-hop questions, which indicates that the reconstructed context is not limited to the directly matched cues, but can incorporate additional evidence from semantically related or temporally neighboring turns. For open-domain questions, CueMem performs slightly worse mainly because these questions require integrating dialogue information with external knowledge, whereas CueMem primarily retrieves and reconstructs evidence from the dialogue history itself, suggesting that explicit external-knowledge retrieval or higher-level memory abstraction could further improve its performance on such questions.

Table~\ref{tab:longmemeval-results} reports the results on LongMemEval. CueMem achieves the best overall accuracy, outperforming the strongest baseline by 1.60\%. This indicates that cue-guided retrieval and dialogue-context reconstruction remain effective even in substantially longer conversation histories. 
For single-session questions, CueMem shows strong performance, achieving the best results on both single-user and single-preference categories and remaining competitive on single-assistant questions. This suggests that fine-grained memory cues are effective for capturing user-specific facts and preferences within localized interaction contexts. Since single-session questions are common in practical conversational agent scenarios, accurately answering such questions is important for improving the user experience and maintaining coherent personalized interactions. CueMem also achieves the best performance on knowledge-update questions, even though it does not introduce a specialized \textsc{UPDATE} operation. This may be because CueMem leverages fine-grained cues to accurately and comprehensively reconstruct query-relevant evidence from the original dialogue while preserving temporal order. Even when the reconstructed context contains conflicting historical information, modern LLMs can often identify and ignore outdated evidence when properly instructed. This result suggests that explicit memory rewriting may not always be necessary for handling knowledge updates. When relevant dialogue evidence is accurately reconstructed and temporally ordered, a sufficiently capable LLM can often resolve conflicts between old and updated information during answer generation.

Across both datasets, CueMem demonstrates consistently strong performance across different long dialogue QA settings. In contrast, most baselines tend to perform well on only one dataset. For example, LightMem achieves stronger results on LongMemEval than on LoCoMo, while several other baselines obtain reasonable performance on LoCoMo but degrade substantially on LongMemEval. This contrast suggests that CueMem is less sensitive to dataset-specific characteristics and can generalize more consistently across conversations with different lengths, structures, and question distributions.

\subsection{Ablation Study}

In this section, we conduct two ablation studies on LoCoMo and LongMemEval to evaluate the effectiveness of turn graph expansion and the priority instruction, respectively.

On LoCoMo, we disable graph expansion while keeping the same cue extraction and retrieval components. The ablated variant therefore answers questions using only the anchor turns associated with the retrieved cues as context. As shown in Figure~\ref{fig:ablation}(a), removing the turn graph reduces the overall accuracy from 81.1\% to 71.4\%, with the largest drops on single-hop and multi-hop questions. This suggests that the retrieved anchor turns alone often do not contain sufficient evidence, and that graph expansion is important for recovering temporally adjacent and semantically related supporting turns.

As discussed above, CueMem does not perform an explicit \textsc{UPDATE} operation, yet it still achieves strong performance on knowledge-update questions. We attribute this result largely to the priority instruction used during answer generation. In practice, this is implemented by adding a simple prompt instruction: ``Prioritize the information most recent and closest to the question time''. To verify whether this simple instruction is indeed effective, we remove it while keeping the retrieved evidence context unchanged. As shown in Figure~\ref{fig:ablation}(b), removing the priority instruction lowers the overall accuracy from 75.2\% to 72.4\%, and the accuracy on knowledge-update questions drops from 87.2\% to 80.8\%. In contrast, several non-update categories remain unchanged or show only minor differences. Temporal questions also show a noticeable accuracy drop of 6.8\%, indicating that the priority instruction not only helps the model identify updated facts, but also guides it toward more accurate temporal reasoning. These ablation results suggest that memory systems can handle knowledge updates without relying on complex update operations. Instead, when relevant evidence is accurately reconstructed, the inherent reasoning capability of the LLM, combined with a simple priority instruction, can effectively resolve update-sensitive questions.

\begin{table}[t]
\centering
\small
\resizebox{\columnwidth}{!}{
\begin{tabular}{llcccc}
\toprule
\multirow{2}{*}{Dataset} & \multirow{2}{*}{Method}
& \multirow{2}{*}{Tokens}
& \multicolumn{2}{c}{Time (s)}
& \multirow{2}{*}{Acc. (\%)} \\
\cmidrule(lr){4-5}
& & & Recon. & Infer. & \\
\midrule
\multirow{2}{*}{LoCoMo}
& Full History & $\sim$21K & -- & 5.95 & 80.19 \\
& CueMem & $\sim$2K & 0.13 & 3.29 & 81.10 \\
\midrule
\multirow{2}{*}{LongMemEval}
& Full History & $\sim$108K & -- & 29.64 & 42.80 \\
& CueMem & $\sim$2K & 0.15 & 3.58 & 70.00 \\
\bottomrule
\end{tabular}
}
\caption{Efficiency and effectiveness comparison between the full-history LLM setting and CueMem. All numbers are averaged per query. Tokens denote the number of input tokens provided to the answer-generation LLM. Recon. denotes the query-time context reconstruction cost, including query encoding, cue retrieval, and graph expansion. Infer. denotes the LLM answer-generation latency.}
\label{tab:efficiency}
\end{table}

\subsection{Efficiency Analysis}
As LLMs continue to support increasingly long context windows, directly feeding the entire dialogue history into the LLM has become a simple and seemingly attractive solution for long-term conversational question answering.
In this section, we compare CueMem with the full-history LLM setting in terms of both effectiveness and query-time efficiency, focusing on answer accuracy, token consumption, and answering latency, with CueMem using a reconstructed context of approximately 2K tokens.

Table~\ref{tab:efficiency} shows that CueMem is substantially more efficient than the full-history LLM setting while achieving better accuracy. On LoCoMo, CueMem consumes only about 10\% of the input tokens used by the full-history setting while achieving higher accuracy. Meanwhile, it reduces the overall query latency by 42.5\%. On LongMemEval, this advantage becomes even more pronounced. Compared with the full-history setting, CueMem uses less than 2\% of the input tokens, achieves a 27.2\% absolute improvement in accuracy, and reduces the overall query latency by 87.4\%.

These results suggest that long-context LLMs may work reasonably well with moderately long histories, but can degrade when the input approaches the maximum context window (such as 128K limit for Llama-3.3). This degradation may be caused by excessive irrelevant context, noise, or the lost-in-the-middle effect. Moreover, the full-history strategy inevitably increases inference cost and latency as the conversation grows. In contrast, CueMem mitigates these issues with only a small context reconstruction overhead, substantially reducing token consumption and latency while improving answer accuracy. This low overhead is enabled by CueMem's fine-grained cues, whose atomic semantics can be effectively encoded by a lightweight 22M-parameter embedding model (all-MiniLM-L6-v2) with 384-dimensional representations, making the retrieval module lightweight enough for resource-constrained settings.

\section{Conclusion}

We presented CueMem, a lightweight and effective long-term memory framework for conversational agents. CueMem follows a simple cue-to-anchor-to-context pipeline: it treats extracted memories as fine-grained retrieval cues rather than self-contained evidence, links them to source turns, and reconstructs query-relevant dialogue context through expansion over a turn graph. This design keeps memory management simple while retaining direct links to the original dialogue evidence. Experiments on LoCoMo and LongMemEval show that CueMem consistently outperforms representative long-term memory baselines while reducing query-time token consumption and latency compared with the full-history setting. These results suggest that cue-guided context reconstruction offers a simple and scalable strategy for long-term memory in conversational agents.

\bibliography{ref}

\end{document}